\documentclass[journal]{IEEEtran}

\usepackage{amsmath}
\usepackage{amssymb}
\usepackage{bm}
\usepackage{mathtools}
\usepackage{verbatim}
\usepackage{tablefootnote}

\usepackage{algorithm}
\usepackage{algorithmic}

\def\etal{\emph{et al}.}

\def\ie{\emph{i.e.}}

\usepackage{bbold}
\usepackage{bbm}
\usepackage{booktabs} 
\usepackage{array}  %
\usepackage[tight,normalsize,sf,SF]{subfigure}
\usepackage{multirow}
\usepackage{url}
\usepackage{color}
\usepackage{float}
\usepackage{gensymb}
\usepackage{cite}

\usepackage{tabu}
\usepackage{dcolumn}

\ifCLASSINFOpdf
\else
\fi
\usepackage{graphicx}

\usepackage[pagebackref=true,breaklinks=true,letterpaper=true,colorlinks,bookmarks=false]{hyperref}

\begin{document}
%
\title{PRA-Net: Point Relation-Aware Network for 3D Point Cloud Analysis}

%
\author{Silin Cheng$^{\ast}$,
        Xiwu Chen$^{\ast}$,
        Xinwei He,
        Zhe Liu,
        and Xiang Bai$^{\dagger}$,~\IEEEmembership{Senior Member,~IEEE}

\thanks{$^{\ast}$Authors contribute equally.}
\thanks{$^{\dagger}$Corresponding author.}
\thanks{This work was supported by the National Program for Support of Top-notch Young Professionals and the Program for HUST Academic Frontier Youth Team 2017QYTD08.}
\thanks{S. Cheng, X. Chen, Z. Liu, and X. He are with the School of Electronic Information and Communications, Huazhong University of Science and Technology, Wuhan, 430074, China. E-mail: \{slcheng, xiwuchen, zheliu1994\}@hust.edu.cn, eriche.hust@gmail.com.}
\thanks{X. Bai is with the School of Artificial Intelligence and Automation, Huazhong University of Science and Technology, Wuhan, 430074, China. E-mail: xbai@hust.edu.cn.}

}
\maketitle

\begin{abstract}
Learning intra-region contexts and inter-region relations are two effective strategies to strengthen
feature representations for point cloud analysis. However, unifying the two strategies for point cloud representation is not fully emphasized in existing methods. To this end, we propose a novel framework named Point Relation-Aware Network (PRA-Net), which is composed of an Intra-region Structure Learning (ISL) module and an Inter-region Relation Learning (IRL) module. The ISL module can dynamically integrate the local structural information into the point features, while the IRL module captures inter-region relations adaptively and efficiently via a differentiable region partition scheme and a representative point-based strategy. Extensive experiments on several 3D benchmarks covering shape classification, keypoint estimation, and part segmentation have verified the effectiveness and the generalization ability of PRA-Net. Code will be available at \url{https://github.com/XiwuChen/PRA-Net}.
\end{abstract}

\begin{IEEEkeywords}
point cloud, shape analysis, intra-region contexts, inter-region relations
\end{IEEEkeywords}

%
\IEEEpeerreviewmaketitle

\section{Introduction}\label{sec:introduction}

\IEEEPARstart{W}{ITH} the proliferation of 3D sensing devices, 3D point cloud analysis has attracted increasing attention due to its generalized applications in numerous prospective fields such as autonomous driving~\cite{wang2019pseudo}, robotics~\cite{ahmed2018edge}. However, this task remains a great challenge since point clouds are normally irregular and non-uniform. In light of the great success of deep learning on 2D images, there have been many efforts to extend the techniques
for 3D point clouds. A straightforward way is to transform point clouds into regular voxel grids~\cite{qi2016volumetric,gadelha2018multiresolution,meng2019vv} or multi-view images~\cite{su2015multi,tatarchenko2018tangent,han20193d2seqviews}. Nonetheless, these methods generally suffer from high computational complexity, large memory overhead, or loss of geometric information. 
As a pioneering work, PointNet~\cite{qi2017pointnet} applies shared pointwise multi-layer perceptrons (MLPs) followed by max-pooling to directly process point clouds. However, it has the following drawbacks that limit its ability to learn more discriminative representations for point clouds: (1) lack of full exploration on the intra-region contextual information. (2) The inter-region relations are also not sufficiently captured. 

Recently, several works focus on deriving the intra-region contexts of the point cloud. For instance, PointNet++~\cite{qi2017pointnet++} introduces a hierarchical framework to capture local contexts by sampling and grouping point clouds. However, since it treats each point individually at local regions and simply uses max-pooling layers to aggregate local features, the geometric relationships among points are not yet fully exploited. To better capture the local structures, DGCNN~\cite{wang2019dynamic} proposes a novel EdgeConv to extract local features by linearly aggregating each point features with its $k$-nearest neighbors. PointWeb~\cite{zhao2019pointweb} considers all the point pairs inside each region and aggregates local features by weighting all edges among points based on feature differences. Nonetheless, the densely connected graph formed in this way may introduce redundant information and lead to high computational overhead. Besides, these methods focus on exploring intra-region contexts while not sufficiently exploit the inter-region relations. 

However, the inter-region relations also provide beneficial information for 3D understanding, such as the similarity, proximity, and symmetry among different parts of an object, which has been found to be important in human visual perception~\cite{wertheimer1923laws}. Several works thus have been proposed to exploit the inter-region relations to further enhance the capacity of the learned 3D shape representations. For example, SPG~\cite{landrieu2018large} breaks down point clouds into geometrically simple parts in an unsupervised manner, where each part is assumed to be semantically homogeneous and assigned with a hard label.
However, it may misclassify those points with different labels but having been partitioned into the same parts. 
LRC-Net~\cite{liu2020lrc} encodes the inter-region context information based on spatial similarity measures, which is inversely proportional to the Euclidean distance among points. It indicates that nearer regions contribute more than the farther regions and suppress the impact of distant regions, which limits the capacity and flexibility of the model to capture long-range inter-region relations.


In this paper, we argue that both the intra-region contexts and inter-region relations should be combined to complement each other and produce better feature representations for 3D point cloud analysis. To this end, we propose a novel end-to-end framework named Point Relation-Aware Network (PRA-Net) to simultaneously exploit the intra-region contexts and inter-region relations. It consists of two core modules: the Intra-region Structure Learning (ISL) module and the Inter-region Relation Learning (IRL) module. The ISL module is employed to capture intra-region contextual information, while the IRL module is designed for exploring inter-region relations. Specifically, for the ISL module, it adopts a novel adaptive fusion mechanism to integrate the local structural information into the point features, thus producing more discriminative intra-region contextual features. For the IRL module, it first dynamically divides the point cloud into a set of regions and then samples one representative point per region. Based on the sampled points, we can easily use them as proxies to model the inter-region relations in an efficient way.
This procedure may repeat multiple times in order to give a more comprehensive evaluation of the inter-region relations. The two modules are interleaved multiple layers to form our final framework, in order to closely and densely capture both the intra-region contexts and inter-region relations. 
The validity of our method has been verified by conducting extensive experiments on four public benchmark datasets, \ie, ModelNet40, ScanObjectNN, KeypointNet, and ShapeNet Part.

In summary, our key contributions are threefold:
\begin{itemize}
 \item We propose an \textbf{I}ntra-region \textbf{S}tructure \textbf{L}earning (ISL) module with a novel fusion mechanism, to adaptively incorporate the local structure information into each point feature, thereby producing more discriminative intra-region contextual information for point clouds.
 \item We present an \textbf{I}nter-region \textbf{R}elation \textbf{L}earning (IRL) module to effectively capture inter-region relations of point clouds. This module dynamically partitions a point cloud into several local regions and samples points from each local region to serve as representative proxies, which can learn the inter-region relations both efficiently and effectively. 
 \item Based on the proposed ISL and IRL modules, we establish an end-to-end framework PRA-Net for point cloud analysis, which has demonstrated its effectiveness and generality on various 3D benchmark datasets involving 3D object classification, keypoint estimation, and part segmentation.
\end{itemize}

The rest of the paper is organized as follows: we review some related works in Section~\ref{sec:related_work} and present the details of the proposed method in Section~\ref{sec:method}. Section~\ref{sec:experiments} provides the details of the experimental results and analysis. Finally, we draw conclusions in Section~\ref{sec:conclusion}.

\section{Related Work} \label{sec:related_work}
\textbf{View-based method.} Given the superiority of the image-based deep learning techniques, many methods~\cite{su2015multi,bai2016gift,han2018seqviews2seqlabels,han20193d2seqviews,he2019view, feng2018gvcnn, wei2020view} project 3D objects into a set of 2D views and learn feature representations from these view images. For instance, MVCNN~\cite{su2015multi} applies a shared 2D CNN to extract view-wise features and simply maxpools
them into a global descriptor. However, the max-pooling operation may lead to information loss by only retaining the maximum values for each view. To address this issue, GVCNN~\cite{feng2018gvcnn} introduces a hierarchical view-group-shape framework to model the correlation in multiple views, and improves the recognition accuracy accordingly. Yang \etal~\cite{yang2019learning} proposes a Relation Network to effectively capture the region-to-region and view-to-view relationships, and achieves better performance. View-GCN~\cite{wei2020view} constructs a view-graph to model the relationship among multiple views and adopts a graph convolutional network to learn the global shape descriptor. However, those approaches are restricted mainly by occlusion surfaces and rendering techniques. 

\textbf{Voxel-based method.} Another way is to voxelize 3D objects into voxels. VoxNet~\cite{maturana2015voxnet} converts input points into voxels and utilizes a 3D CNN to extract its features. However, due to the limitation of sparsity and high computational overhead, some space partition methods are proposed to relieve this issue. OctNet~\cite{riegler2017octnet} uses a set of unbalanced octrees to hierarchically partition the space, which reduces the number of empty voxels and thus is both memory and computation efficient. 
O-CNN presents an octree-based convolutional neural network for 3D shape analysis, whose inputs are the average normal vectors of a 3D model sampled in the finest leaf octants, resulting in compact memory footprint and efficient computation overhead.
Submanifold sparse convolution~\cite{graham20183d} introduces a sparse convolutional operation to reduces the computation overhead, which strictly operates on submanifolds. 

\begin{figure}
\begin{center}
    \includegraphics[width=0.9\linewidth]{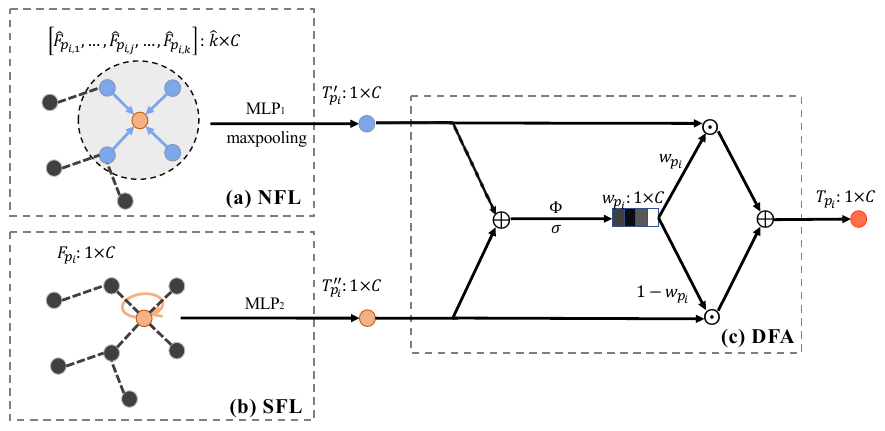}
\end{center}
\caption{\textbf{Intra-region Structure Learning (ISL) module}, which includes \textbf{(a)} a NFL module, \textbf{(b)} a SFL module, and \textbf{(c)} a DFA module. The NFL module is designed to encode local structural information, the SFL module is responsible for capturing global shape information and the DFA module is present for adaptively combining local structural information and global shape information. $F_{{p}_{i}}$ denotes the point feature. $\hat{F}_{{p}_{i,j}}$ represents the edge feature between point $p_{i}$ and its $j$-th nearest neighbor $p_{i,j}$. $T'_{p_{i}}$ and $T''_{p_{i}}$ encode the global shape information and local structural information of point $p_{i}$, respectively. $T_{p_{i}}$ is the output of the ISL module. $C$ is the number of feature channels. $\sigma$ indicates the sigmoid function. $\phi$ is implemented by two fully-connected layers with batch-normalization. $w_{p_{i}}$ is the attention vector to weight $T'_{p_{i}}$ and $T''_{p_{i}}$.  $\oplus$ and $\odot$ are the element-wise summation and Hadamard product, respectively.
\label{Fig.LFE_Compare}}
\end{figure}

\textbf{Point-based method.}
Recently, there exist many works~\cite{qi2017pointnet,qi2017pointnet++,han2020point2node,xu2020geometry,lin2020fpconv,xu2020grid} attempting to directly process 3D point clouds.  PointNet~\cite{qi2017pointnet} performs shared MLPs and a symmetric function to handle the unordered point clouds. 
However, the intra-region structure information is not fully exploited. To solve this problem, many endeavors have been made to effectively aggregate local features.
PointNet++~\cite{qi2017pointnet++} applies the PointNet recursively on each partitioned region and aggregates local features by max-pooling. However, since it treats each point independently, the geometric relationships among points are not fully explored. Motivated by the success of graph neural networks in dealing with irregular data, several works apply it to capture local geometric topologies. DGCNN~\cite{wang2019dynamic} employs an EdgeConv operator to capture local contexts by linearly aggregating center point features with edge features. SpiderCNN~\cite{xu2018spidercnn} introduces a novel filter defined by a product of a step function and a Taylor polynomial.
Grid-GCN~\cite{xu2020grid} adopts a novel grid-based data structuring strategy and aggregates the local structure information by graph convolution. 
In addition, some methods utilize kernel points to capture local structure information. For instance, KCNet~\cite{shen2018mining} leverages a kernel correlation layer and a graph pooling layer to further exploit the geometric feature. InterpCNN~\cite{mao2019interpolated} utilizes a set of discrete kernel weights
and an interpolation function to model geometric relationships between points and kernel-weight coordinates. To overcome the geometric and scale variance problem, KPConv~\cite{thomas2019kpconv} and 3DGCN~\cite{lin2020convolution} adopt deformable kernels to capture local structure information of the point cloud.
RSCNN~\cite{liu2019relation}, PointConv~\cite{wu2019pointconv},  and PCCN~\cite{wang2018deep} obtain the dynamic kernel weights from low-level geometric information (\ie, xyz coordinates), which are then used to aggregate intra-region contexts.

Unlike these methods, our approach adaptively incorporates the local structure information into each point feature, thus can produce more discriminative intra-region features. Furthermore, these methods mainly focus on exploiting local information while not fully explore the inter-region relations. 

Besides, several works further enhance the learned representations by modeling the inter-region relations.
SRN~\cite{duan2019structural} employs shared MLPs to learn structural relational features between different local regions.
LRC-Net~\cite{liu2020lrc} encodes the inter-region context information based on spatial similarity measures, which is inversely proportional to the Euclidean distance among points. A-SCN~\cite{xie2018attentional}, PointASNL~\cite{yan2020pointasnl}, and Point2Node~\cite{han2020point2node} use a global attention mechanism to aggregate features by considering relationships among all the points, which may lead to redundancy in both computation and representation. Instead of using self-attention as an auxiliary module, PointTransformer~\cite{zhao2020point} and PCT~\cite{guo2020pct} design a more general framework which is based on Transformer~\cite{vaswani2017attention} for 3D point cloud processing.
Different from these methods, we dynamically partition the point cloud into different regions and utilize representative points within each region as proxies to model the inter-region relations, which is more efficient and effective.

\section{Proposed Method}\label{sec:method} 

The proposed PRA-Net consists of two key elements. The first is an Intra-region Structure Learning (ISL) module, which tries to enhance each point feature using its corresponding contextual region. The second is a novel Inter-region Relation Learning (IRL) module, which aims at modeling the relations across different local regions. The two modules are combined together to drive our PRA-Net to learn more discriminative representations from point clouds. 

\subsection{Intra-region Structure Learning}
\label{lef_module}

Let $\mathcal{P} = \{p_i\}_{i=1}^{N}$ represents a point cloud of $N$ points with the associated point-wise features $\mathcal{F} = \{F_{{p}_{i}}\}_{i=1}^{N}$, where $F_{{p}_{i}} \in \mathbb{R}^{1 \times C}$ denotes the $C$-dimentional feature of point $p_i$.
For a given sampled point $p_i$, we use its $\hat{k}$ nearest neighbors $\mathcal{N}(p_i) = \{p_{i,j}\}_{j=1}^{\hat{k}}$ to model the local region\footnote{Note that here we assume a point is its 1st nearest neighbor to itself for convenience (i.e.,~$p_{i,1} = p_i$).}. 
The goal of the ISL module is to obtain more discriminative representations of local regions.
Concretely, the computation of ISL can be decomposed into three stages: the Neighbor-based Feature Learning (NFL) stage, the Self-based Feature Learning (SFL) stage, and the Dynamic Feature Aggregation (DFA) stage.

\noindent\textbf{NFL.} Exploring the local geometric structure is essential for point cloud understanding. 
Inspired by EdgeConv~\cite{wang2019dynamic}, we obtain the edge feature $\hat{F}_{{p}_{i,j}} \in \mathbb{R}^{1 \times C}$ in a local coordinate system by computing the difference between the point feature $F_{{p}_{i}}$ and its $j$-th nearest neighbor $F_{{p}_{i,j}}$ as follows $\hat{F}_{{p}_{i,j}} = F_{{p}_{i}}-F_{{p}_{i,j}}$, where $j \in \{1, 2, ..., \hat{k}\}$. Then a shared MLP denoted by $\mathtt{MLP_{1}(\cdot)}$ is applied to encode the local structural information, which can be formulated as 

\begin{equation}
    F^{'}_{{p}_{i,j}} = \mathtt{MLP_{1}}(\hat{F}_{{p}_{i,j}}) \text{,}
\end{equation}
where $F'_{{p}_{i,j}} \in \mathbb{R}^{1 \times C}$ and $C$ is the number of output channels. Finally, we use a max-pooling layer, denoted by $\mathtt{MaxPooling(\cdot)}$, to further aggregate local structural information into a compact one, which is written as 
\begin{equation}
T'_{{p}_{i}} = \mathtt{MaxPooling}([F'_{{p}_{i,1}};...;F'_{{p}_{i,j}};...;F'_{{p}_{i,{\hat{k}}}}]) \text{,}
\label{eq:neighboring-based feature}
\end{equation} where $T'_{{p}_{i}} \in \mathbb{R}^ {1 \times C}$ encodes the local structural information of the region centered at $p_{i}$. 

\noindent\textbf{SFL.} Since the point feature $F_{p_{i}}$ is represented in the global coordinate system, it contains the global shape information of the object \cite{wang2019dynamic}. As shown in Fig.~\ref{Fig.LFE_Compare}~(b), we obtain the point feature $T^{''}_{{p}_{i}} \in \mathbb{R}^{1 \times C}$ by employing another MLP denoted by $\mathtt{MLP_{2}(\cdot)}$ to extract point features as follows
\begin{equation}
    T^{''}_{{p}_{i}} = \mathtt{MLP_{2}}(F_{{p}_{i}}) \text{.}
\label{eq:self-based feature}
\end{equation}

\begin{figure*}
\centering
\includegraphics[width=0.85\linewidth]{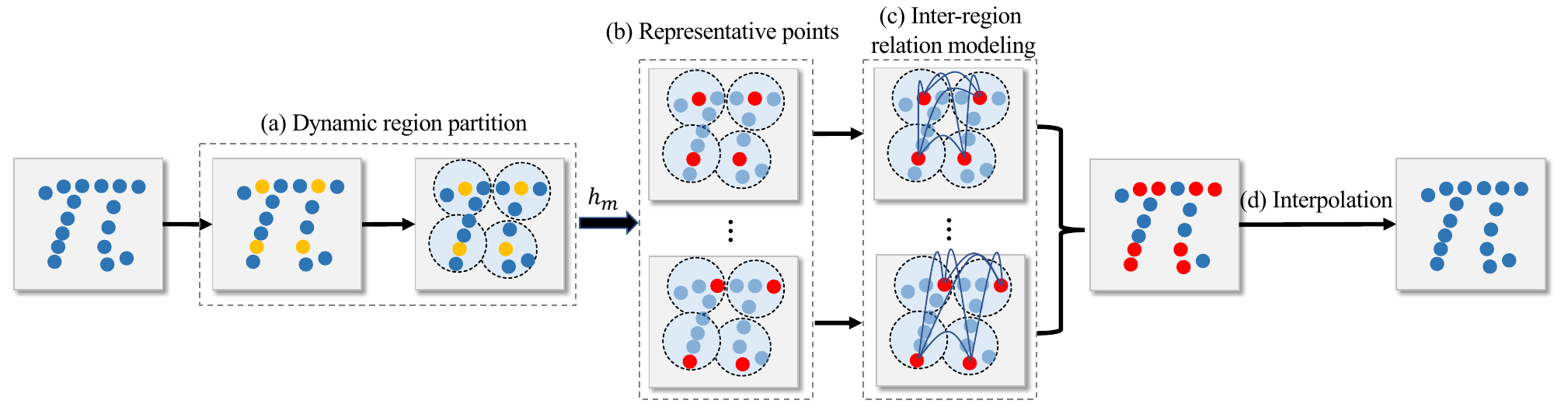}
\caption{\textbf{Inter-region Relation Learning (IRL) module.} This module contains four stages, including dynamic region partition, sampling representative points, inter-region relation modeling, and feature interpolation. \textbf{(a)} Dynamic region partition. Regions are constructed based on the dynamic sampled points (yellow points). \textbf{(b)} Sampling representative points (red points).  \textbf{(c)} Inter-region relation modeling. \textbf{(d)} Feature interpolation.}

\label{Fig.IRESchematicDiagram}
\vspace{-2ex}
\end{figure*}

\noindent\textbf{DFA.} To exploit both local structure information captured by $T'_{{p}_{i}}$ and global shape information captured by $T''_{{p}_{i}}$, EdgeConv~\cite{wang2019dynamic} utilizes linear aggregation (\ie, addition) to integrate them. However, linear aggregation is not feature-adaptive and thus limits the capacity of the learned point features. To better leverage the local structural information and global shape information, we propose a dynamic fusion mechanism to achieve this.
As shown in Fig.\ref{Fig.LFE_Compare}~(c), we first fuse $T^{'}_{{p}_{i}}$ and $T^{''}_{{p}_{i}}$ via an element-wise summation to form a compact feature representation, which is then compressed into an attention vector $\mathbf{w}_{p_{i}} \in \mathbb{R}^ {1 \times C}$ as follows

\begin{equation}
\mathbf{w}_{p_{i}} = \sigma(\Phi(T'_{p_{i}} + T''_{p_{i}})) \text{,}
\end{equation}
where $\sigma(\cdot)$ represents the sigmoid activation function, $\Phi(\cdot)$ is the transformation network which is implemented by two fully-connected layers with batch normalization. 
Then the final feature representation $T_{p_i} \in \mathbb{R}^{1 \times C}$ of point $p_{i}$ is computed as follows

\begin{equation}
T_{p_{i}} = \mathbf{w}_{p_{i}} \odot T^{'}_{p_i}+ (\boldsymbol{1} - \mathbf{w}_{p_{i}}) \odot T^{''}_{p_i} \text{,}
\end{equation}
where $\odot$ denotes the Hadamard product and $\boldsymbol{1} \in \mathbb{R}^{1 \times C}$ is an all ones-vector.

\subsection{Inter-region Relation Learning}

The IRL module aims at enhancing point features by modeling inter-region relations. It first adaptively decomposes the point cloud into a set of local regions, where each local region is composed of $k$-nearest neighbors of a center point. 
Concretely, it first selects $S$ points $\{p_1, p_2,..., p_S\}$ as centroids via a dynamic region partition method and then applies the $k$-NN algorithm to construct the corresponding regions for each sampled center point, which are denoted by
$\{\mathcal{N}(p_1), \mathcal{N}(p_2), ..., \mathcal{N}(p_S)\}$. In this way, we obtain $S$ local regions with each containing $k$ neighbor points, as shown in Fig.~\ref{Fig.IRESchematicDiagram}~(a). 
Lastly, the proposed IRL module tries to ``connect'' these local regions directly to model their relations. 

\begin{figure}
\begin{center}
    \includegraphics[width=0.8\linewidth]{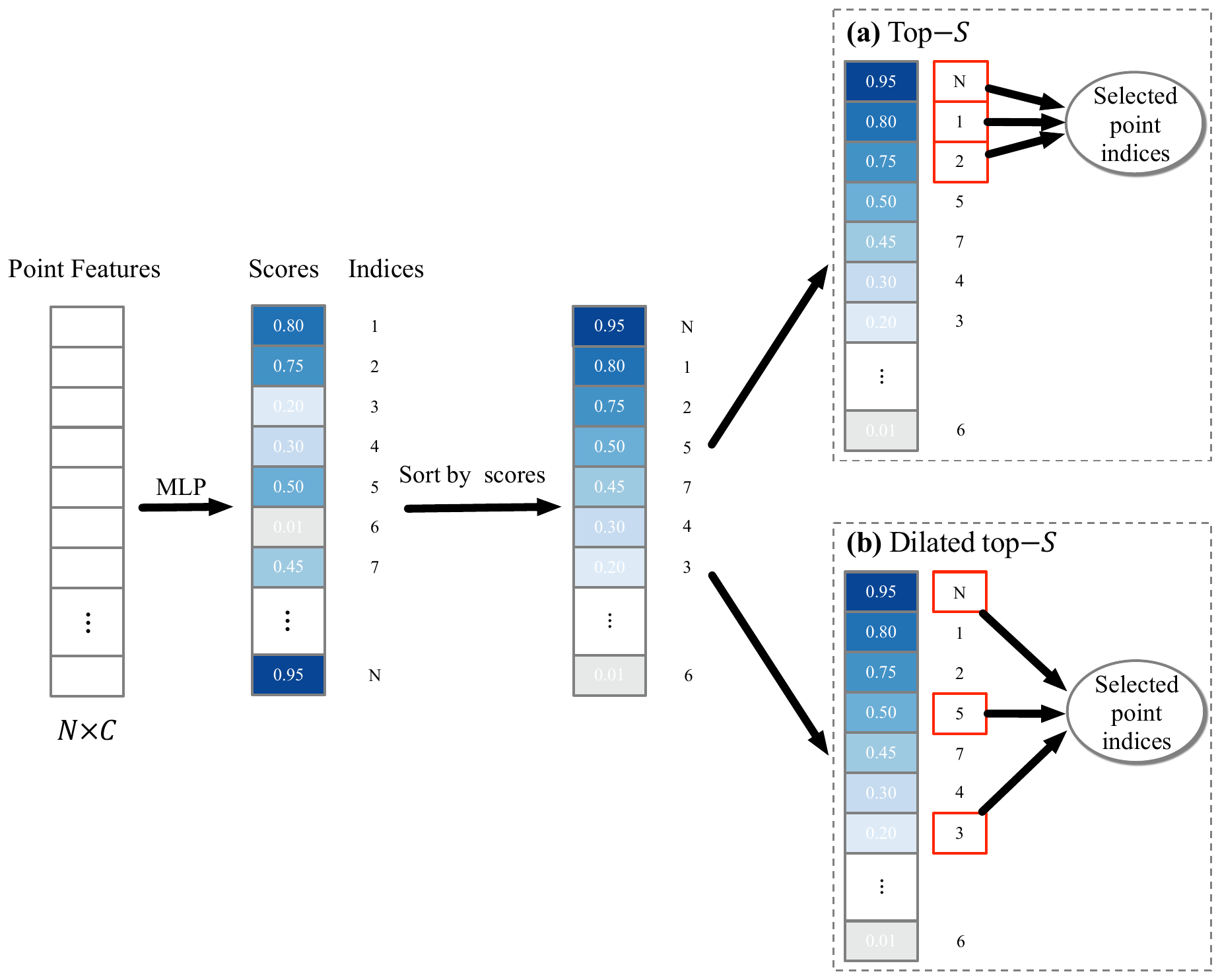}
\end{center}
\caption{\textbf{Illustration of the dynamic region partition stage.} \textbf{(a)} The top branch is the top-$S$ partition method, \textbf{(b)} the bottom branch is the dilated top-$S$ partition method. The numbers in the red rectangles represent the selected point indices. For simplicity, we set $S=3$ here for visualization. (Best viewed in color.)}
\label{Fig.DynamicSampling}
\end{figure}

\begin{figure}
\begin{center}
    \includegraphics[width=0.8\linewidth]{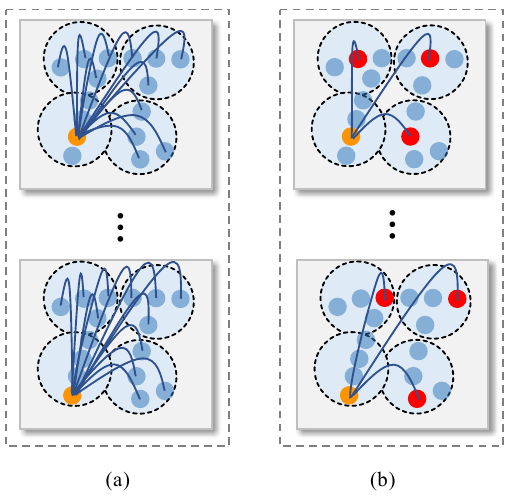}
\end{center}
\caption{\textbf{Two different ways to model inter-region relations.} \textbf{(a)} The naive version approach. The orange point features are enhanced by all the points (blue points) from other regions. \textbf{(b)} The representative point-based version. The point features (orange color) are enhanced by the sampled representative points (red points) from other regions.}
\label{Fig.naive}
\end{figure}

\noindent\textbf{Dynamic Region Partition.} A simple choice is adopting the Farthest Point Sampling (FPS) algorithm to sample key points and use their $k$-nearest neighbors to construct local regions. However, FPS is heuristic and task-irrelevant, which limits its ability to generalize to various scenarios. Based on this observation, we propose a novel region partition strategy that is capable of partitioning the regions adaptively in a data-driven way.  
Specifically, we first employ a shared MLP to project $T_{p_{i}}$ into a score $s_{{p}_{i}}$ as
\begin{equation}
    s_{p_{i}} = \sigma(\mathtt{MLP_{3}}(T_{p_{i}})) \text{,}
\end{equation}
where 
$s_{{p}_{i}}$ is the score which implies the importance of the point $p_{i}$.
Then we select $S$ points from the point cloud according to the calculated importance scores. 
One straightforward approach is to sample the top-$S$ points as centroids based on the scores to construct local regions, as shown in Fig.~\ref{Fig.DynamicSampling}~(a). Nonetheless, considering that points close to each other tend to get similar scores, it may select a bunch of points clustered around a small region, while suppresses those points that also contain discriminative information but scatter at other parts of the input point cloud. To solve this problem, we propose a dilated top-$S$ partition method to keep a balance between the importance and the diffusion of points during the region partition process. Concretely, instead of taking the top-rank points as centroids, we sample $S$ points with stride $N/S$, thus make the selected region evenly distributed on the input point cloud, as shown in Fig.~\ref{Fig.DynamicSampling}~(b). 

Finally, instead of adopting a discrete region partition strategy, we perform a scalar multiplication between the score $s_{p_{i}}$ of each center point $p_{i}$ and the feature $T_{p_{i,j}}$ of $p_{i}$'s neighbors to make this process differentiable as 
\begin{equation}
 G_{{p}_{i,j}} = s_{p_{i}}T_{p_{i,j}}\text{,}
\end{equation}
where $G_{{p}_{i,j}} \in \mathbb{R}^{1 \times C}$ is updated point features,  $p_{i,j} \in \mathcal{N}(p_i)$, and $j$ $\in \{1, 2, 3,..., k\}$.


\noindent\textbf{Inter-region Relations Learning.} In order to learn the inter-region relations, one naive approach is to enhance each point's feature using all the points from other regions. As shown in Fig.~\ref{Fig.naive}~(a), in order to update the selected point feature (i.e.,~the yellow point), one has to connect all the points from other regions to it. In this case, the number of edges equals $(S-1)k^2$ per region. Thus the total number of edges is $S(S-1)k^2$. 
However, this approach relies on constructing a very dense graph to achieve this, which is intractable to use in practice. To overcome the issues incurred by the above naive approach, we improve upon it by sampling representative points from each region and then use self-attention to model the inter-region relations, as shown in Fig.~\ref{Fig.naive}~(b). Since a sampled point from a local region can give us some cues about a region, thus learning relations about the sampled $S$ points (i.e.,~one point per region) can give us a hint on the inter-region relations. Such a strategy not only endows our method with efficiency but also improves the representation learning by sparsifying the connections among different regions in a balanced manner. Fig.~\ref{Fig.IRESchematicDiagram} gives an overview of the IRL module. Specifically, it consists of the following three stages:

\textbf{1)~Sampling Representative Points.} It is the key step in our methods. Let us define $h_m$ as the sampling strategy that takes a local region $\mathcal{N}(p_i)$ as input, and outputs $m$ representative points, which is written as

\begin{equation}
    \chi_{i} =h_m(\mathcal{N}(p_i))= [r_{i,1}, ..., r_{i,t}, ..., r_{i,m}]\text{,}\label{eq:h}
\end{equation} where $r_{i,t} \in \mathcal{N}(p_i)$ represents the $t$-th representative point for the local region $\mathcal{N}(p_i)$. A natural choice of $h_m$ is to use a random sampling method. In light of the randomness introduced by random sampling, we adopt a stable sampling strategy named $k$-NN based method. The $k$-NN based method samples $m$ representative points one by one from each region in ascending order based on their distances to the center point.
Moreover, we also explore using max-pooling and mean-pooling methods to aggregate features to represent each local region. 
The different choices of $h_m$ are analyzed in Sec.~\ref{table:sample_strategy}. Such a sampling process will be applied to all the regions in the point cloud. Then we pack representative points of the same order in each $\chi_{i}$, which gives us $\chi = [\chi_{0};\chi_{1}; ...; \chi_{S}]^\mathrm{T}$. It should be noted that each row in $\chi$ represents representative points from each region, which can be seen as an aspect of the original point cloud. 

\begin{figure*}[t]
\begin{center}
  \includegraphics[width=0.95\linewidth]{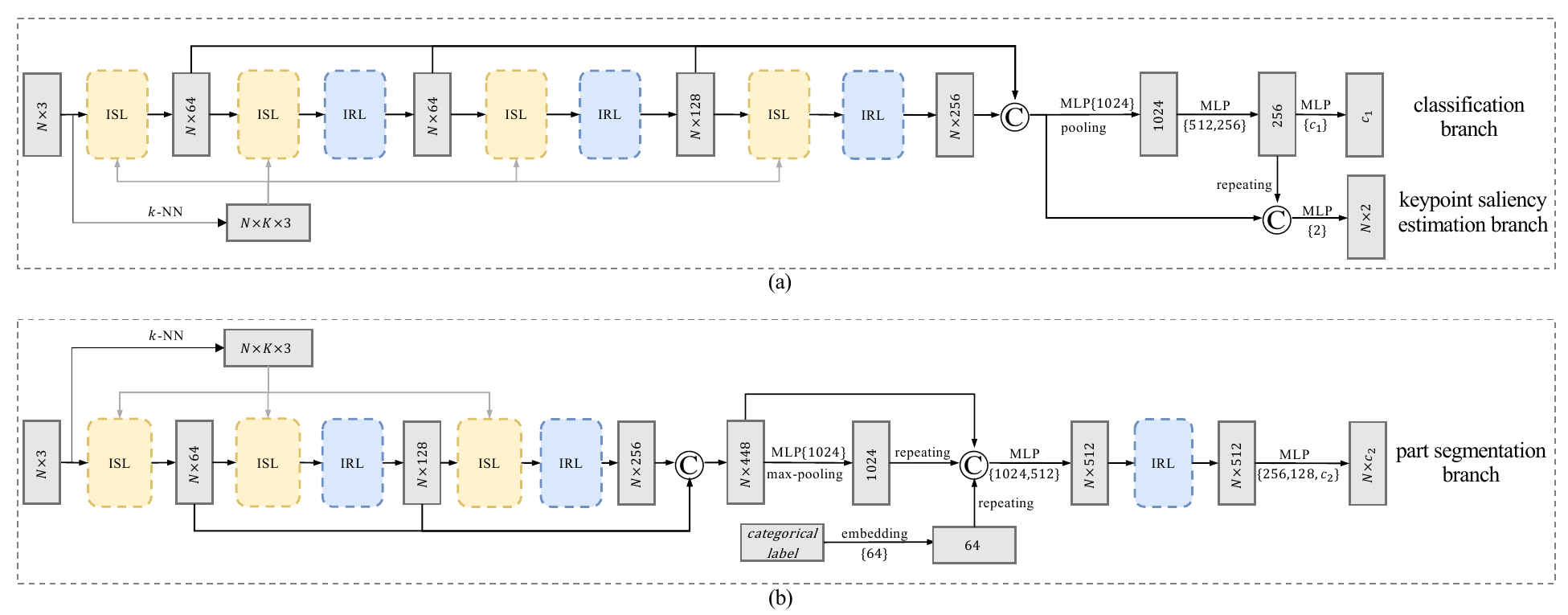}
\end{center}
\caption{\textbf{Network architectures.} \textbf{(a)} The architecture designed for classification (the top branch) and keypoint saliency estimation (the bottom branch) tasks. \textbf{(b)} The architecture designed for part segmentation task. They all contain two components: the ISL module and the IRL module. $N$ is the number of input points. $c_1$ and $c_2$ are the number of classification and part segmentation classes, respectively. ``MLP" stands for a multi-layer perceptron. $\copyright$: concatenation. } 
\label{Fig.PR-Net}
\end{figure*}

\textbf{2)~Inter-region Relation Learning.} After we have obtained $\chi$, we then perform self-attention~\cite{vaswani2017attention} over the features corresponding to each row, i.e.,~$G_t=\{G_{r_{1,t}}, G_{r_{2,t}},..., G_{r_{S,t}}\}$, where $G_{r_{i,t}}$ represents the feature of point $r_{i,t}$ obtained in the dynamic region partition stage.
With self-attention, the inter-region relations can be easily captured. Mathematically, it is formulated as 
\begin{equation}
\label{eq:IRE}
    \hat{G}_{r_{i,t}} = W_{z} \sum_{j=1}^{S} \frac{f(G_{r_{i,t}}, G_{r_{j,t}})}{\mathnormal{Z(G)}}(W_{v} \cdot G_{r_{j,t}})\text{,}
\end{equation}
where $W_{z}$ and $W_{v}$ represent linear transform matrices.  We denote $w_{ijt}=\frac{f(G_{r_{i,t}}, G_{r_{j,t}})}{\mathnormal{Z(G)}}$ as a normalized similarity score (or attention score) of the query point $r_{i,t}$ and the key point $r_{j,t}$, where $f(G_{r_{i,t}}, G_{r_{j,t}})$ denotes the relationship between point $r_{i,t}$ and $r_{j,t}$ and ${Z(G)}$ is the normalization factor. As softmax is a commonly used normalization technique, we use it to normalize $w_{ijt}$, which can be written as
\begin{equation}
\label{eq:output_feature}
\begin{split}
    w_{ijt} =   \frac{exp(\langle W_{q} \cdot G_{r_{i,t}}, W_{k} \cdot G_{r_{j,t}}\rangle)}{\sum_{l=1}^{S} exp(\langle W_{q} \cdot G_{r_{i,t}} , W_{k} \cdot G_{r_{l,t}}\rangle)}\text{,}
\end{split}
\end{equation}
where $W_{q}$ and $W_{k}$ denote linear transform matrices corresponding to the query point and key point, respectively.

One superiority of our method is that we can reduce the computational complexity to some extent compared with the naive method. In particular, for the naive version, we have to build a complete graph as each pair of points from different regions is ``connected" by an edge. In summary, this heavily-connected graph contains $S(S-1)k^{2}$ edges in total. While our improved representative-point based attention only needs to build a very sparse graph to model the inter-region relations. In this case, the number of edges is $(S-1)m$ per region, and the total number of edges is reduced to $S(S-1)m$, which is just $\frac{m}{k^2}$ of the naive implementation. To be more specific, if $m$ equals 4 and $k$ equals 8, the number of connections in the graph would become approximately $1/16$ of the naive approach. 

\textbf{3)~Feature Interpolation.}  
For the IRL module, we further add a residual link in order to alleviate the information loss and accelerate the learning process. However, since the output is of different size to the original input, we apply feature interpolation to interpolate it back to the input space of the IRL module.
Similar to PointNet++, we obtain the interpolated feature $\Delta G_{p_{v}}$ of $p_{v}$ by utilizing an inverse Euclidean distance weighted average method based on its 3-nearest neighbors $\{p_u\}_{u=1}^{3}$ in $\chi$ as

\begin{equation}
    \Delta G_{p_{v}} =\frac{\sum_{u=1}^{3} \lambda_{u}(p_{u})\hat{G}_{p_{u}}}{\sum_{u=1}^{3}\lambda_{u}(p_{u})} \text{,}
\end{equation}
where $v \in \{1, 2,  ..., N\}$, $\lambda_{u}(p_{u})= \frac{1}{{d(p_{v}, p_{u})}^{2}}$ is an inverse $L2$ distance weight. Then the interpolated feature will be added to the original points' embeddings to generate the final output of the IRL module.

\subsection{Network Architecture Details} 


By combining both the ISL module and IRL module, we propose three different frameworks for the classification, keypoint estimation, and segmentation tasks, respectively. 

\noindent\textbf{Classification Network.} As shown in Fig.~\ref{Fig.PR-Net}~(a) (the top branch), our classification framework consists of four LFE and three IRL modules. For the input layer, the first LFE module is used to extract the low-level local features. While for the remaining layers, we insert LFE and IRL modules alternately to learn more informative representations for each point by considering both the intra-region contextual information and inter-region relations.
Shortcut connections are further used to aggregate richer representations via concatenating the outputs of the first LFE module and all the IRL modules. A shared fully-connected layer is then used to obtain the global feature. Inspired by~\cite{woo2018cbam}, we concatenate max-pooling and mean-pooling outputs to get a better representation. Finally, we feed it into three fully-connected layers to make the final prediction.

\noindent\textbf{Keypoint Saliency Estimation Network.} The keypoint saliency estimation network is similar to the classification network, which is shown in Fig.~\ref{Fig.PR-Net}~(a) (the bottom branch). Following \cite{you2020keypointnet}, we obtain the pointwise features by concatenating the output of the first LFE module, all the IRL modules, and the penultimate layer of the network. Lastly, we use them to estimate keypoint saliency for each point.

\noindent\textbf{Segmentation Network.} The part segmentation network is shown in Fig.~\ref{Fig.PR-Net}~(b). Similar to the classification network, three ISL modules and two IRL modules are interleaved to capture the intra-region contexts and inter-region relations. Then a shared MLP and a max-pooling layer are further utilized to learn the global feature. In addition, we also use an embedding layer to encode the object label and apply a shortcut connection to fuse it with all previous outputs to get a more discriminative representation. Finally, another IRL module is used, which is followed by three fully-connected layers for segmentation. 

\subsection{Network Configuration Details}\label{details}

\noindent\textbf{Architecture protocol.}
In this section, we detail the parameter configurations of the Intra-region Structural Learning (ISL) module and the Inter-region Relation Learning (IRL) module in classification network, keypoint saliency estimation network, and part segmentation network, respectively. For clarity, we first introduce several notations as follows:

ISL($\hat{k}$, [$l_{1}$;...;$l_{d}$]): the ISL module, where $\hat{k}$ is the number of the nearest neighbors of each point in the local region, [$l_{1}$;...;$l_{d}$] represents shared MLPs of $d$ layers with width $l_{i}$ ($i$=1,...,$d$).

IRL($S$, $k$, $m$): the IRL module, where $S$ is the number of regions, $k$ and $m$ represent the total number of points and the number of representative points in each local region, respectively. 

Then, we list the configuration details of the ISL and IRL modules from left to right in Fig.\ref{Fig.PR-Net}.

\noindent\textbf{Classification and Keypoint Saliency Estimation Network}\label{classification details}. 
ISL(20, [64]) $\rightarrow$ ISL(20, [64]) $\rightarrow$ IRL(256, 4, 4) $\rightarrow$ ISL(20, [128]) $\rightarrow$ IRL(128, 8, 4) $\rightarrow$ ISL(20, [256]) $\rightarrow$ IRL(64, 16, 4).

\noindent\textbf{Part Segmentation Network}. ISL(32, [64, 64, 64]) $\rightarrow$ ISL(32, [128, 128, 128])  $\rightarrow$ IRL(128, 16, 8) $\rightarrow$  ISL(32, [256, 256, 256]) $\rightarrow$ IRL(256, 16, 8) $\rightarrow$ IRL(128, 32, 16).

\section{Experiments} \label{sec:experiments}

To demonstrate the effectiveness of the proposed PRA-Net, we conduct our experiments on various tasks such as shape classification, keypoint saliency estimation, and part segmentation.

\subsection{Point Cloud Classification}

\noindent\textbf{Datasets.} We evaluate our method on two types of public datasets: synthetic dataset, \ie, ModelNet40~\cite{wu20153d}, and real-world dataset, \ie, ScanObjectNN~\cite{uy2019revisiting}. 
ModelNet40 consists of 12,311 3D synthetic objects in 40 categories, with 9,843 models for training and 2,468 models for testing. ScanObjectNN is a challenging real-world object dataset which is built upon the SceneNN~\cite{hua2016scenenn} and ScanNet~\cite{dai2017scannet} datasets. It includes 15,000 objects in 15 classes. 

\noindent\textbf{Implementation Details.} For experiments on the ModelNet40 dataset, we sample 1,024 points for each 3D object following~\cite{qi2017pointnet}. 
We use the $k$-NN based strategy to sample representative points. Data augmentation techniques including random scaling and translation are utilized. The random scaling factor ranges from 0.66 to 1.33 and the translation ranges from \mbox{-0.2} to 0.2. For optimization, we use SGD to train our model for 250 epochs with a batch size of 32. The initial learning rate is set to 0.1 and the cosine annealing method is used to dynamically adjust the learning rate. To alleviate the overfitting problem, the label smoothing technique~\cite{Szegedy_2016_CVPR} is used during training. For ScanObjectNN, we conduct experiments on its hardest variant, \ie, PB\_T50\_RS as stated in ~\cite{uy2019revisiting}, which contains various perturbations including translation that randomly shifts the bounding box up to 50\% of its size, rotation ,and scaling on each object.
We use the same data augmentation techniques following~\cite{uy2019revisiting}. The model and training strategy are the same as those on the ModelNet40. The metrics are overall accuracy (OA) and average class accuracy (mAcc).

\noindent\textbf{Synthetic object classification.} We compare our method with representative state-of-the-art methods on the ModelNet40 dataset and the results are summarized in Table~\ref{tab:comparsions on cls}.
As shown, our method achieves 90.5\% and 93.2\% in terms of mAcc and OA, respectively, which outperforms almost all the state-of-the-art methods using only 1,024 points as input. 
Since our method is point-cloud based, we mainly compare our method with previous point cloud based methods. 
Compared with InterpCNN~\cite{mao2019interpolated} and DGCNN~\cite{wang2019dynamic}, which concentrate on aggregating intra-region contexts, we improve upon them by 0.2\% and 0.3\% in OA, respectively. 
In addition, for those methods focusing on how to effectively integrate intra-region contexts and inter-region relations of the point cloud, we also obtain competitive results over them. Specifically, when compared with PointASNL~\cite{yan2020pointasnl}, which adopts 5-time voting during testing, we still outperform it by 0.3\%.  It should also be mentioned that compared with KPConv~\cite{li2018so} uses 6$k$ points as input, while our method uses only 1$k$ points as input. However, we still achieve a superior result (92.9\% \emph{vs.} 93.2\% in OA). What's more, when using 2k points as the input, the pro-posed PRA-Net can achieve the best performance among all reported results. All these results consistently demonstrate the competitiveness of our method. 

\noindent\textbf{Real-world object classification.} The ScanObjectNN dataset introduces great challenges to existing point cloud classification techniques due to the existence of background noise, missing parts, and diverse deformations. The classification results are presented in Table~\ref{tab:comparsions on cls}. As shown, when using only 1,024 points as input, our approach outperforms all the other point-based methods by a large margin. Particularly, it significantly surpasses the second-best point-based method, \ie, PointCNN~\cite{li2018pointcnn} with an increase of 2.5\% in OA and 2.8\% in mAcc. Moreover, compared with the recent DGCNN~\cite{wang2019dynamic}, we obtain 2.8\% improvements in terms of OA. To investigate the impact of the number of input points on PRA-Net, we also train the network with 2048 points and obtain an impressive accuracy of 82.1\% in OA. The superior performance on the real-world dataset indicates that our method has great potential in practical applications.

\begin{table}[!tb]
\small
\centering
\caption{Classification results~(\%) on the ModelNet40 and ScanObjectNN datasets (``-": unknown, nor.: normal).}
\label{tab:comparsions on cls}
\vspace{-2ex}
\begin{tabular}{|l|p{1cm}<{\centering}|*{2}{p{0.75cm}<{\centering}}|*{2}{p{0.75cm}<{\centering}}|}
\hline
\multirow{2}{*}{Methods}& \multirow{2}{*}{Inputs} & \multicolumn{2}{c|}{ModelNet40} & \multicolumn{2}{c|}{ScanObjectNN}  \\
\cline{3-4} \cline{5-6} 
    &   & mAcc & OA  & mAcc & OA  \\
\hline
\hline
Subvolume~\cite{qi2016volumetric} & voxels &- &89.2&-&-\\
OctNet~\cite{riegler2017octnet} & octree &83.8& 86.5&-&-\\
ECC~\cite{simonovsky2017dynamic} & graphs &82.4& 87.0&-&-\\
\hline
\hline
KPConv~\cite{thomas2019kpconv} & 6k &-&92.9&-&-\\
PointConv~\cite{wu2019pointconv} & 1k, nor. &- & 92.5&-&-\\
SO-Net~\cite{li2018so} & 2k, nor. & 87.3 & 90.9&-&-\\
Point2Cap~\cite{wen2020point2spatialcapsule} & 1k, nor. & - & \textbf{93.7} & - & - \\
\hline
\hline
$\psi$-CNN~\cite{lei2019octree} & 1k & 88.7& 92.0 &-&-\\
PAT~\cite{yang2019modeling} & 1k & - & 91.7 & - & -\\
RSCNN\tablefootnote[2]{For fair comparisons, we report the results of RSCNN and DensePoint without voting strategy according to their paper.\label{voting}}
~\cite{liu2019relation} & 1k &-&92.9&-&-\\
DensePoint\textsuperscript{\ref {voting}}~\cite{liu2019densepoint} & 1k & -&92.8&-&-\\
A-SCN~\cite{xie2018attentional} & 1k & 87.6&90.0 &-&-\\
InterpCNN~\cite{mao2019interpolated} & 1k &-&93.0&-&-\\
FPConv~\cite{lin2020fpconv} & 1k &-&92.5&-&-\\
3D-GCN~\cite{lin2020convolution} &1k &-&92.1&-&-\\
PointASNL~\cite{yan2020pointasnl} & 1k & - & 92.9&-&-\\
Point2Seq~\cite{liu2019point2sequence} & 1k & 90.4 & 92.6 &- &- \\
MAP-VAE~\cite{han2019multi} & 1k &-& 90.2 &-& -\\
A-CNN~\cite{komarichev2019cnn} & 1k &90.3& 92.6 &-& -\\
Point2Node~\cite{han2020point2node} & 1k &- & 93.0 &-& -\\
Point2Cap~\cite{wen2020point2spatialcapsule} & 1k & - & 93.4 & - & - \\
PosPool~\cite{liu2020closer}& 1k & - & 93.2 & - & - \\
FatNet~\cite{94db77733d1a457180565c42fbfa6301}& 1k & 90.6 & 93.2 & - & - \\
SpiderCNN~\cite{xu2018spidercnn} & 1k &- & - &69.8&73.7\\
3DmFV~\cite{ben20183dmfv} & 1k &- &91.4& 58.1 & 63.0\\
PointNet++~\cite{qi2017pointnet++}   & 1k &-&90.7&75.4&77.9\\
PointNet~\cite{qi2017pointnet}     & 1k &86.2 &89.2 &63.4 &68.2\\
DGCNN~\cite{wang2019dynamic} & 1k &90.2& 92.9 & 73.6& 78.2\\
PointCNN~\cite{li2018pointcnn} & 1k & 88.1& 92.2 & 75.1 & 78.5\\
\hline
\hline
PRA-Net (\textbf{Ours}) & 1k &90.6& 93.2 & 77.9 & 81.0\\

PRA-Net (\textbf{Ours}) & 2k & \textbf{91.2} & \textbf{93.7} & \textbf{79.1} & \textbf{82.1} \\
\hline
\end{tabular}
\end{table}

\subsection{Keypoint Saliency Estimation}

\begin{table*}[t]
\small
\centering
\caption{Keypoint saliency estimation results in mIoU~(\%) with a distance threshold of 0.01 on the KeypointNet benchmark~(``-": unknown).}
\label{table:KeypointNet_mIoU}
\begin{tabular}{|l|*{1}p{0.96cm}<{\centering}|
*{1}p{0.45cm}<{\centering}*{1}p{0.45cm}<{\centering}*{1}p{0.45cm}<{\centering}*{1}p{0.45cm}<{\centering}
*{1}p{0.45cm}<{\centering}*{1}p{0.45cm}<{\centering}*{1}p{0.45cm}<{\centering}*{1}p{0.45cm}<{\centering}
*{1}p{0.45cm}<{\centering}*{1}p{0.45cm}<{\centering}*{1}p{0.45cm}<{\centering}*{1}p{0.45cm}<{\centering}
*{1}p{0.45cm}<{\centering}*{1}p{0.45cm}<{\centering}*{1}p{0.45cm}<{\centering}*{1}p{0.45cm}<{\centering}
|}

\hline
Methods & mIoU & air. & bat. & bed & bot. & cap & car & cha. & gui. & hel. & kni. & lap. & mot & mug & ska. & tab. & ves.\\
\hline
\hline
RSCNN~\cite{liu2019relation}& 15.9 & 21.0& 11.9& 19.3& 11.6& 18.9& 15.8& 17.6& 17.9& 0.0& 24.2& 25.3& 13.4& 17.2& 5.9& 23.7& 10.1 \\
SpiderCNN~\cite{xu2018spidercnn} & 11.7 & 22.2 & 7.2 & 17.7 & 4.1 & 2.7 & 5.5 & 15.9 & 7.1 & 0.0 & \textbf{30.0} & 22.4 & 14.5 & 4.9 & 0.0 & 23.9 & 8.5 \\
PointNet++~\cite{qi2017pointnet++} & 12.3 & 13.1 & 8.3 & 16.3 & 11.1 & 19.9 & 13.0 & 12.6 & 9.5 & 2.1 & 18.5 & 19.3 & 16.3 & 9.2 & 5.6 & 14.0 & 8.0\\
PointNet~\cite{qi2017pointnet} & 2.8 & 9.1 & 0.5 & 6.4 & 0.0 & 0.0 & 0.0 & 4.5 & 0.0 & 0.0 & 0.0 & 11.6 & 1.9 & 0.0 & 0.0 & 11.0 & 0.0\\
PointConv~\cite{wu2019pointconv} & 17.9 & 25.3 & 15.2 & \textbf{32.4} & 7.3 & 13.5 & 20.3 & 21.7 & 21.2 & 2.1 & 5.0 & 27.8 & 18.9 & 21.7 & 13.2 & 26.8 & 13.9\\
RSNet~\cite{huang2018recurrent} & 15.1 & 20.5 & 12.8 & 19.2 & 13.1 & 15.7 & 15.1 & 13.9 & 16.4 & 8.4 & 18.3 & 22.8 & 20.2 & 16.8 & 4.0 & 15.4 & 9.7\\
DGCNN~\cite{wang2019dynamic} & 20.8 & 32.3 & 17.7 & 21.6 & 15.0 & \textbf{21.5} & 15.1 & 23.8 & 20.7 & 3.5 & 29.4 & 30.1 & 23.5 & 18.1 & 12.8 & 31.7 & 15.6\\
\hline
PRA-Net (\textbf{Ours}) & \textbf{28.6} & \textbf{38.6} & \textbf{28.1} & 25.8 & \textbf{37.1} & 13.4 & \textbf{31.4} & \textbf{27.2} & \textbf{32.3} & \textbf{14.2} & 29.4 & \textbf{44.1} & \textbf{26.7} & \textbf{26.3} & \textbf{22.8} & \textbf{37.6} & \textbf{21.7}  \\
\hline
\end{tabular}
\end{table*}

\begin{table*}[t]
\small
\centering
\caption{Keypoint saliency estimation results in mAP~(\%) with a distance threshold of 0.01 on the KeypointNet benchmark~(``-": unknown).}
\label{table:KeypointNet_mAP}
\begin{tabular}{|l|*{1}p{0.96cm}<{\centering}|
*{1}p{0.45cm}<{\centering}*{1}p{0.45cm}<{\centering}*{1}p{0.45cm}<{\centering}*{1}p{0.45cm}<{\centering}
*{1}p{0.45cm}<{\centering}*{1}p{0.45cm}<{\centering}*{1}p{0.45cm}<{\centering}*{1}p{0.45cm}<{\centering}
*{1}p{0.45cm}<{\centering}*{1}p{0.45cm}<{\centering}*{1}p{0.45cm}<{\centering}*{1}p{0.45cm}<{\centering}
*{1}p{0.45cm}<{\centering}*{1}p{0.45cm}<{\centering}*{1}p{0.45cm}<{\centering}*{1}p{0.45cm}<{\centering}
|}

\hline
Methods & mAP & air. & bat. & bed & bot. & cap & car & cha. & gui. & hel. & kni. & lap. & mot & mug & ska. & tab. & ves.\\
\hline
\hline
RSCNN~\cite{liu2019relation}& 23.2 & 34.4 & 17.3 & 28.4 & 16.8 & \textbf{31.6} & 16.3 & 21.7 & 18.2 & 3.1 & 30.6 & 37.9 & 23.8 & 25.4 & 9.7 & 41.4 & 14.7 \\
SpiderCNN~\cite{xu2018spidercnn} & 13.7 & 25.8 & 6.7 & 19.8 & 2.7 & 4.0 & 6.5 & 18.9 & 10.5 & 0.4 & 28.4 & 34.3 & 15.0 & 5.3 & 1.7 & 30.2 & 8.9 \\
PointNet++~\cite{qi2017pointnet++} & 16.5 & 16.9 & 11.4 & 21.9 & 15.2 & 27.7 & 16.7 & 14.5 & 12.6 & 3.4 & 19.7 & 27.0 & 22.0 & 11.4 & 5.7 & 30.5 & 7.8\\
PointNet~\cite{qi2017pointnet} & 5.6 & 8.5 & 3.6 & 6.4 & 1.3 & 3.2 & 2.3 & 9.6 & 1.0 & 0.4 & 16.3 & 14.5 & 2.6 & 3.4 & 1.7 & 12.0 & 2.2\\
PointConv~\cite{wu2019pointconv} & 25.5 & 28.1 & 24.6 & 45.8 & 10.1 & 15.7 & 24.6 & 30.8 & 21.7 & 2.0 & 17.3 & 46.5 & 29.3 & 27.3 & 18.9 & 42.4 & 22.6\\
RSNet~\cite{huang2018recurrent} & 22.0 & 31.1 & 17.8 & 29.5 & 12.8 & 21.8 & 21.8 & 15.4 & 16.1 & 6.1 & 31.5 & 35.0 & 26.1 & 23.2 & 5.4 & 45.3 & 12.8\\
DGCNN~\cite{wang2019dynamic} & 28.8 & 43.8 & 26.2 & 33.4 & 20.7 & 27.6 & 21.4 & 30.3 & 22.9 & 4.8 & \textbf{40.5} & 46.4 & 29.2 & 24.9 & 17.3 & 52.1 & 19.7\\
\hline
PRA-Net (\textbf{Ours}) & \textbf{43.2} & \textbf{47.0} & \textbf{45.2} & \textbf{50.9} & \textbf{61.2} & 17.0 & \textbf{54.5} & \textbf{38.7} & \textbf{43.1} & \textbf{16.4} & 34.7 & \textbf{70.4} & \textbf{43.5} & \textbf{38.5} & \textbf{39.2} & \textbf{61.4} & \textbf{30.0} \\
\hline
\end{tabular}
\end{table*}

\begin{table*}[t]
\small
\centering
\caption{Part segmentation results~(\%) on the ShapeNet Part benchmark~(``-": unknown).}
\label{table:shapenetpart}
\begin{tabular}{|l|*{1}p{0.82cm}<{\centering}*{1}p{0.82cm}<{\centering}|
*{1}p{0.35cm}<{\centering}*{1}p{0.35cm}<{\centering}*{1}p{0.35cm}<{\centering}
*{1}p{0.35cm}<{\centering}*{1}p{0.35cm}<{\centering}*{1}p{0.35cm}<{\centering}
*{1}p{0.35cm}<{\centering}*{1}p{0.35cm}<{\centering}*{1}p{0.35cm}<{\centering}
*{1}p{0.35cm}<{\centering}*{1}p{0.35cm}<{\centering}*{1}p{0.35cm}<{\centering}
*{1}p{0.35cm}<{\centering}*{1}p{0.35cm}<{\centering}*{1}p{0.35cm}<{\centering}
*{1}p{0.4cm}<{\centering}
|}

\hline
Methods & Cls. mIoU & Ins. mIoU & air. & bag & cap & car & cha. & ear. & gui. & kni. & lam. & lap. & mot. & mug & pis. & roc. & ska. & tab.\\
\hline
\hline
PointNet~\cite{qi2017pointnet} & 80.4 & 83.7 & 83.4 & 78.7 & 82.5 & 74.9 & 89.6 & 73.0 & 91.5 & 85.9 & 80.8 & 95.3 & 65.2 & 93.0 & 81.2 & 57.9 & 72.8 & 80.6\\
PointNet++~\cite{qi2017pointnet++} & 81.9 & 85.1 & 82.4 & 79.0 & 87.7 & 77.3 & 90.8 & 71.8 & 91.0 & 85.9 & 83.7 & 95.3 & 71.6 & 94.1 & 81.3 & 58.7 & 76.4 & 82.6\\
SCN~\cite{xie2018attentional} & 81.8 & 84.6 & 83.8 & 80.8 & 83.5 & 79.3 & 90.5 & 69.8 & 91.7 & 86.5 & 82.9 & 96.0 & 69.2 & 93.8 & 82.5 & 62.9 & 74.4 & 80.8\\
SGPN~\cite{wang2018sgpn} & 82.8 & 85.8 & 80.4 & 78.6 & 78.8 & 71.5 & 88.6 & 78.0 & 90.9 & 83.0 & 78.8 & 95.8 & \textbf{77.8} & 93.8 & \textbf{87.4} & 60.1 & \textbf{92.3} & \textbf{89.4}\\
SyncSpecCNN~\cite{yi2017syncspeccnn} & 82.0 & 84.7 & 81.6 & 81.7 & 81.9 & 75.2 & 90.2 & 74.9 & \textbf{93.0} & 86.1 & 84.7 & 95.6 & 66.7 & 92.7 & 81.6 & 60.6 & 82.9 & 82.1\\
RSNet~\cite{huang2018recurrent} & 81.4 & 84.9 & 82.7 & 86.4 & 84.1 & 78.2 & 90.4 & 69.3 & 91.4 & 87.0 & 83.5 & 95.4 & 66.0 & 92.6 & 81.8 & 56.1 & 75.8 & 82.2\\
PointCNN~\cite{li2018pointcnn} & \textbf{84.6} & 86.1 & 84.1 & 86.5 & 86.0 & \textbf{80.8} & 90.6 & 79.7 & 92.3 & \textbf{88.4} & 85.3 & 96.1 & 77.2 & \textbf{95.3} & 84.2 & 64.2 & 80.0 & 83.0\\
DGCNN~\cite{wang2019dynamic} & 82.3 & 85.1 & 84.2 & 83.7 & 84.4 & 77.1 & 90.9 & 78.5 & 91.5 & 87.3 & 82.9 & 96.0 & 67.8 & 93.3 & 82.6 & 59.7 & 75.5 & 82.0\\
Point2Seq~\cite{liu2019point2sequence} & - & 85.2 & 82.6 & 81.8 & 87.5 & 77.3 & 90.8 & 77.1 & 91.1 & 86.9 & 83.9 & 95.7 & 70.8 & 94.6 & 79.3 & 58.1 & 75.2 & 82.8\\
A-CNN~\cite{komarichev2019cnn} & - & 85.9 & 83.9 & 86.7 & 83.5 & 79.5 & 91.3 & 77.0 & 91.5 & 86.0 & 85.0 & 95.5 & 72.6 & 94.9 & 83.8 & 57.8 & 76.6 & 83.0\\
RSCNN~\cite{liu2019relation}&84.0 & 86.2 & 83.5 & 84.8 & 88.8 & 79.6 & 91.2 & 81.1 & 91.6 & \textbf{88.4} & \textbf{86.0} & 96.0 & 73.7 & 94.1 & 83.4 & 60.5 & 77.7 & 83.6\\
DensePoint~\cite{liu2019densepoint} &84.2 & \textbf{86.4} & 84.0 & 85.4 & \textbf{90.0} & 79.2 & 91.1 & \textbf{81.6} & 91.5 & 87.5 & 84.7 & 95.9 & 74.3 & 94.6 & 82.9 & \textbf{64.6} & 76.8 & 83.7\\
DPAM~\cite{liu2019dynamic} & - & 86.1 & 84.3 & 81.6 & 89.1 & 79.5 & 90.9 & 77.5 & 91.8 & 87.0 & 84.5 & \textbf{96.2} & 68.7 & 94.5 & 81.4 & 64.2 & 76.2 & 84.3\\ 
ShellNet~\cite{zhang2019shellnet} & 82.8 & - & 84.3 & 79.6 & 88.9 & 79.1 & 90.0 & 79.4 & 91.3 & 85.9 & 82.3 & 95.4 & 68.6 & 94.9 & 82.7 & 61.5 & 79.7 & 81.7\\
Point2Cap~\cite{wen2020point2spatialcapsule} & - & 85.3 & 83.5 & 83.4 & 88.5 & 77.6 & 90.8 & 79.4 & 90.9 & 86.9 & 84.3 & 95.4 & 71.7 & \textbf{95.3} & 82.6 & 60.6 & 75.3 & 82.5 \\
\hline
PRA-Net (\textbf{Ours}) & 83.7 & 86.3 & \textbf{84.4} & \textbf{86.8} & 89.5 & 78.4 & \textbf{91.4} & 76.4  & 91.5 & 88.2 & 85.3 & 95.7& 73.4& 94.8& 82.1 & 62.3& 75.5 & 84.0 \\

\hline
\end{tabular}
\end{table*}
\noindent\textbf{Dataset.} Keypoint saliency estimation is a challenging task which requires the model to predict whether a point is the keypoint or not. We evaluate the proposed PRA-Net for this task on the KeypointNet dataset. KeypointNet is built on ShapeNetCore~\cite{yi2016scalable}. It contains 8,329 mesh models with 83,231 keypoints from 16 categories. Each model is downsampled to 2,048 points during data preprocessing. The dataset is split into train, validation, and test datasets, with 7:1:2 ratio.

\noindent\textbf{Implementation Details.} In keypoint saliency estimation network, we use the same architecture and hyperparameters as the classification network. We use Adam~\cite{kingma2014adam} to optimize our model and the initial learning rate is set to 0.001. Following ~\cite{you2020keypointnet}, we report mean Intersection over Unions (mIoU) and mean Average Precisions (mAP) with a distance threshold of 0.01.

\begin{figure*}[t]
\begin{center}
  \includegraphics[width=0.8\linewidth]{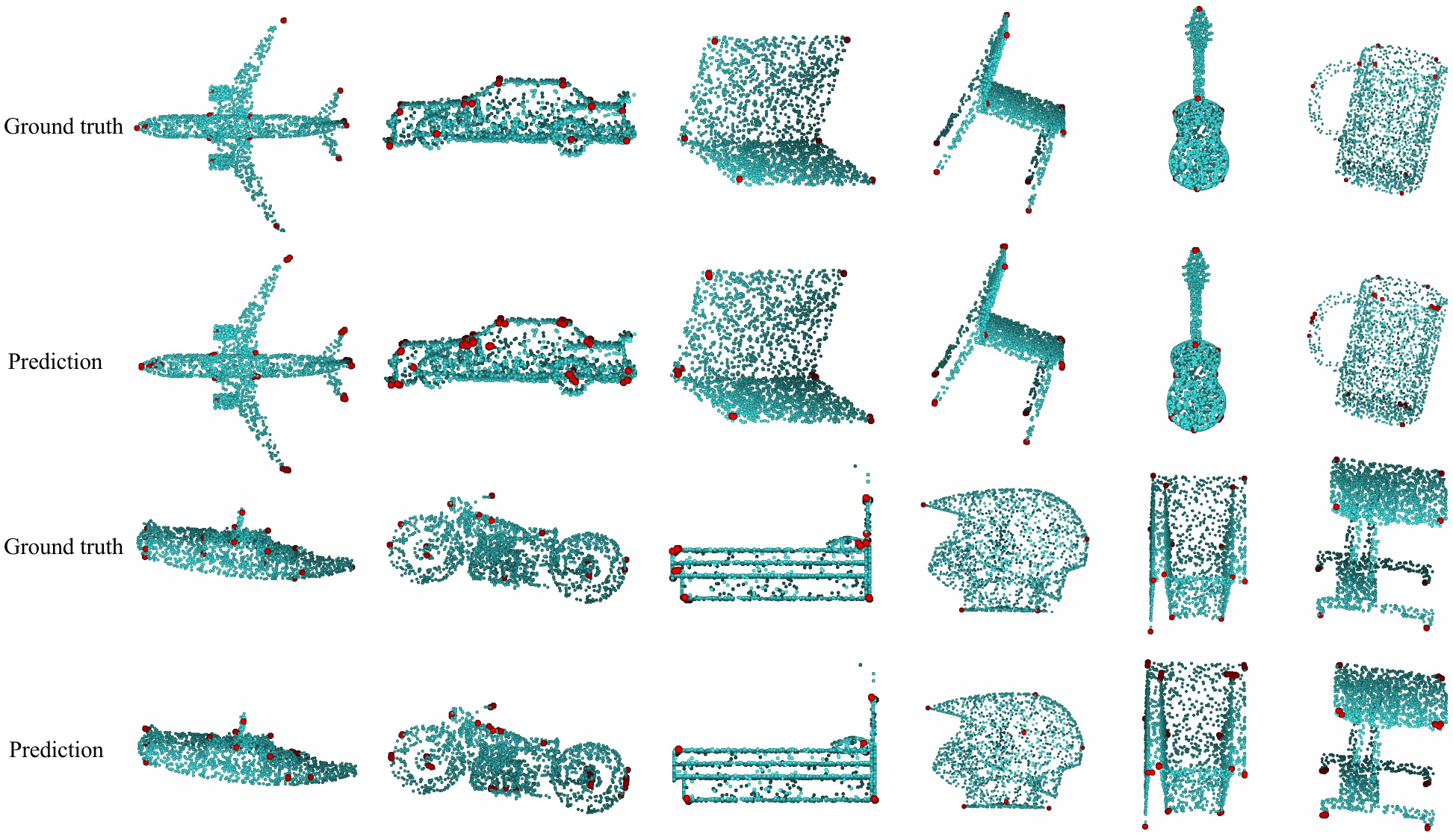}
\end{center}
\caption{The qualitative results on the KeypointNet dataset. The red points are keypoints.}
\label{Fig.KeypointNet visualization result}
\end{figure*}

\noindent\textbf{Results.} As shown
in Table~\ref{table:KeypointNet_mIoU} and Table~\ref{table:KeypointNet_mAP}, our method achieves the best performance with mIoU of 28.6\% and mAP of 43.2\%. Compared with the second-best method DGCNN~\cite{wang2019dynamic}, we improve upon it by 7.8\% and 14.4\% in mIoU and mAP, respectively. Moreover, we also provide the qualitative results shown in Fig.~\ref{Fig.KeypointNet visualization result}. It can be observed that our method can accurately identify the keypoints annotated on each object. These results demonstrate the effectiveness of our method on this challenging dataset. 


\subsection{Part Segmentation}

\noindent\textbf{Dataset.} We further evaluate PRA-Net on part segmentation task. 
ShapeNet Part benchmark~\cite{yi2016scalable} is adopted for the experiment. This dataset consists of 16 shape categories and 50 different part classes. Each object is annotated with one shape category and 2 to 6 part classes. There are 16,880 models in total, with 14,006 for training and 2,874 for testing.

\noindent\textbf{Implementation Details.} In part segmentation network, we set $\hat{k}$ to 32 in each ISL module and set $S$ to 256, 256, 128 in each IRL module respectively. The sampling strategy and data augmentation techniques are the same as the classification task. Following ~\cite{qi2017pointnet++}, we randomly select $2,048$ points of each object to train our model. We use Adam~\cite{kingma2014adam} to optimize our model with a mini-batch size of 16. The initial learning rate is set to 0.001 and decays with a rate of 0.5 every 30 epochs. The initial momentum value of batch normalization is set to 0.9 and reduces by half every 30 epochs.

\begin{figure}[t]
\begin{center}
    \includegraphics[width=0.95\linewidth]{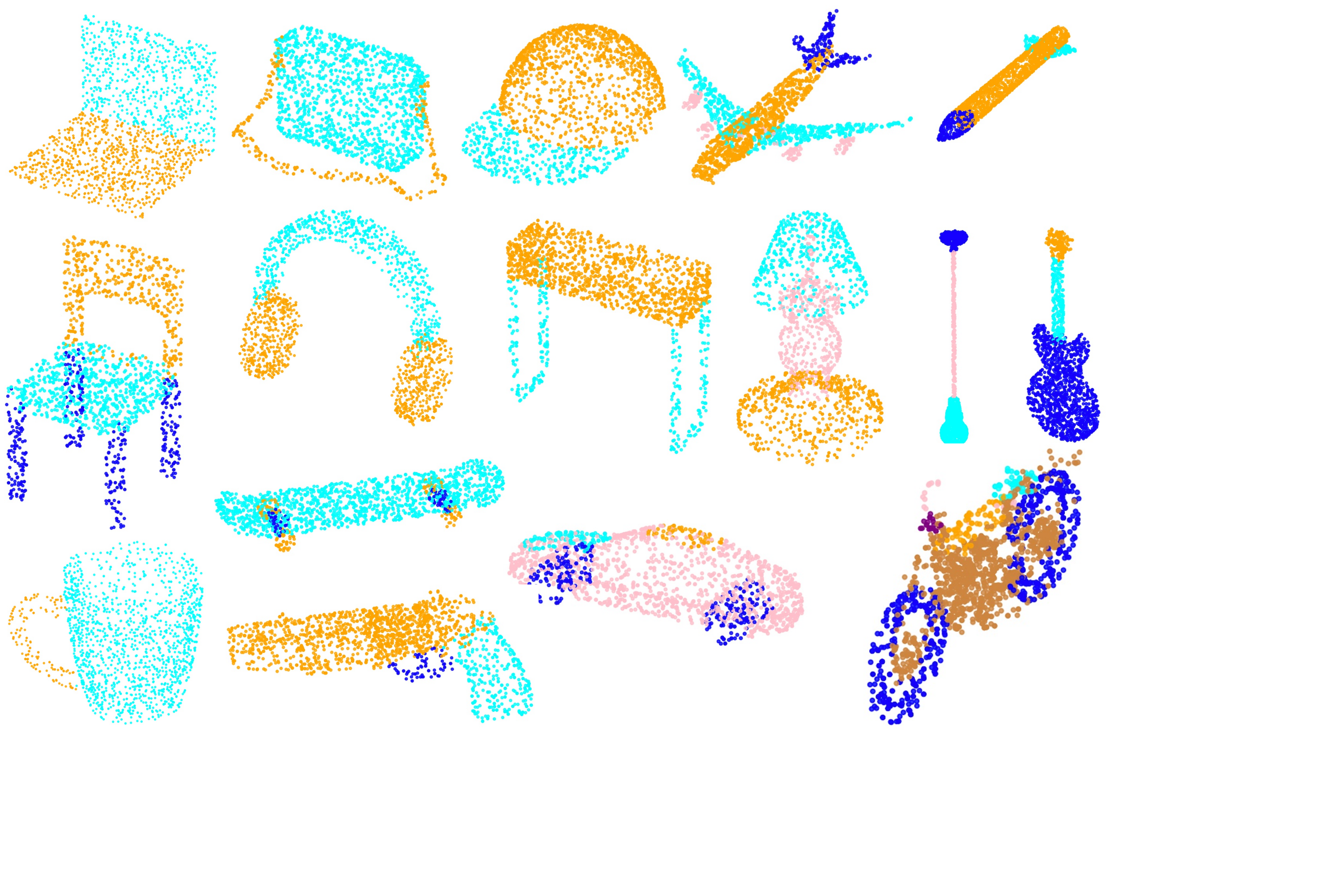}
\end{center}
\caption{The qualitative results on the ShapeNet Part dataset.}
\label{Fig.part segmentation visualization results}
\end{figure}

\noindent\textbf{Results.} Table ~\ref{table:shapenetpart} summarizes quantitative experimental results of various typical methods, where our method achieves competitive values in instance mIoU of $86.3\%$ and class mIoU of $83.7\%$. Concretely, our model outperforms PointNet++~\cite{qi2017pointnet++}, DGCNN~\cite{wang2019dynamic} and A-CNN~\cite{komarichev2019cnn}. When compared with RSCNN~\cite{liu2019relation}, our model shows better performance in instance mIoU and slightly worse performance in class mIoU. The visualization of segmentation results is shown in Fig.~\ref{Fig.part segmentation visualization results}. It can be seen that our model is capable of segmenting out semantic parts of each object faithfully.

\subsection{Discussions}\label{ablation}
To further illustrate the effectiveness of the proposed ISL and IRL module, we design an ablation study on both the shape classification and keypoint saliency estimation tasks.

\begin{figure}[t]
\begin{center}
    \includegraphics[width=0.95\linewidth]{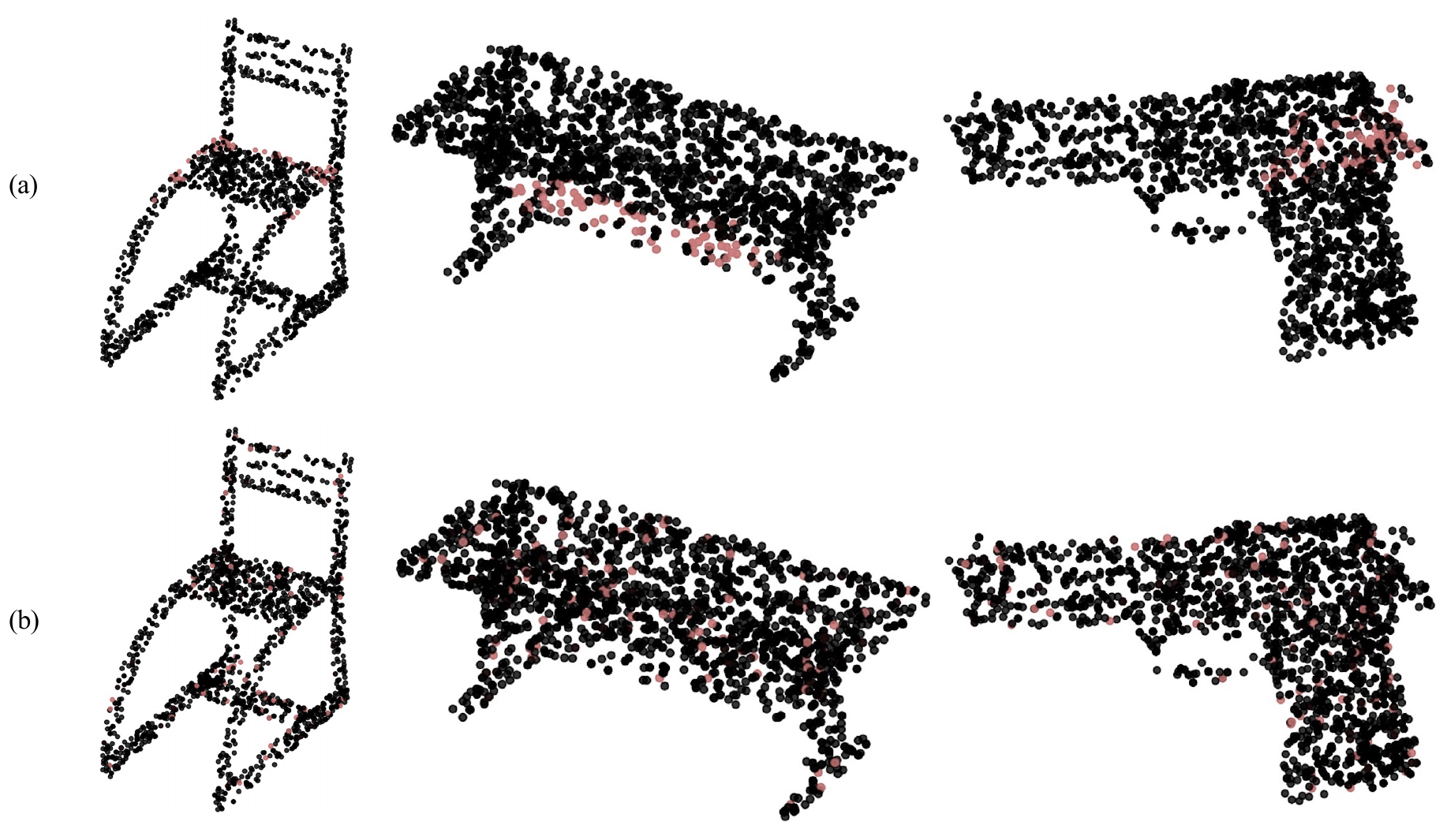}
\end{center}
\caption{The visualization of the sampled points (orange points) by different dynamic region partition methods. (a) Top-$S$ method. (b) Dilated top-$S$ method. Best viewed in color.}
\label{Fig.DynamicRegionPartitionComp}
\end{figure}

\begin{figure*}[t]
\begin{center}
    \includegraphics[width=0.8\linewidth]{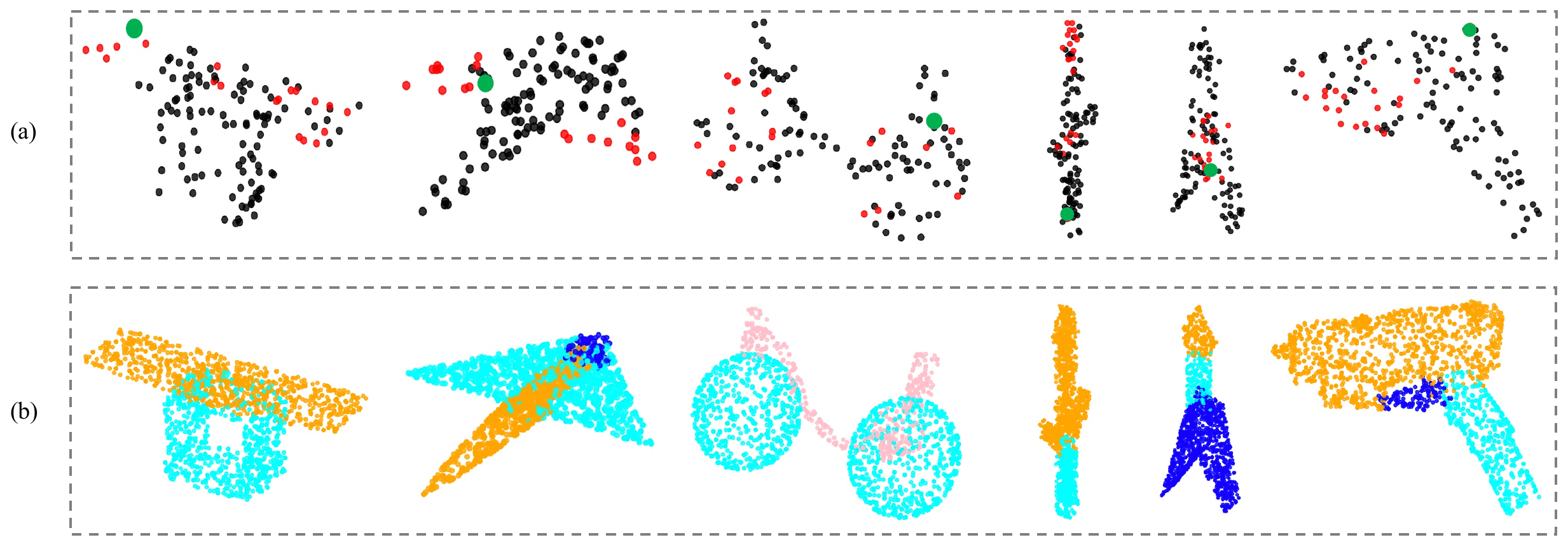}
\end{center}
\caption{\textbf{(a)} The visualization of the top-20 points (red points) that have the most influence on the anchor points (green points) according to the attention weights of the last IRL module on the ShapeNet Part dataset and \textbf{(b)} the corresponding segmentation ground truth.}
\label{Fig.ShapeNetPart Attention Visualization}
\end{figure*}

\noindent\textbf{Effect of Different Components.} The goal of this ablation study is to show the effect of each component in PRA-Net. Results are summarized in Table~\ref{table:effectiveness_of_components}. Here, we set two baselines: model A and model B. Both two modules mainly focus on exploiting the intra-region contextual information by the ISL module. The difference is that model A encodes global shape information by SFL while model B encodes local structural information by NFL. The baseline model A obtains a low classification accuracy of 66.5\% and model B gets 76.6\%, respectively. When we use linear aggregation,~\ie, addition, to combine global shape information and local structural information (model C), an improvement of 2.1\% is obtained. Then, when we replace the linear aggregation of global shape information and local structural information with dynamic feature aggregation (model D), the accuracy can be further improved to 79.8\%. The results demonstrate that our dynamic feature aggregation can better leverage the local structural information and global shape information and produce more discriminative features than the linear aggregation method. Eventually, in order to validate the effectiveness of the IRL module, we add it to model D which gives us model E. It can be seen that with all the components (model E), PRA-Net achieves the best performance. Model E gains a 1.2\% improvement compared to model D, which indicates the importance of learning inter-region relations.


\noindent\textbf{Different Region Partition Methods.} To better understand the influence of different three region partition methods, we conduct the ablation experiments on the ScanObjectNN and KeypointNet dataset. The quantitative results are summarized in Table~\ref{table:effectiveness_of_different_region_partition}. As shown, the proposed dilated top-$S$ method can obtain more discriminative results than the other alternatives. To further conceptualize the difference between top-$S$ and dilated top-$S$ partition methods, we visualize the selected points of the two methods in Fig.~\ref{Fig.DynamicRegionPartitionComp}, respectively. As shown, the points selected by the top-$S$ method are clustered around a small region. However, the points selected by the dilated top-$S$ method are more evenly distributed, thus can better describe the geometric topology of the object and lead to more promising results than the top-$S$ method. Though the FPS can also select points that are evenly scattered over the whole point set, it is task agnostic and not learnable, which limits its representation power.

\noindent\textbf{Different Sampling Strategies $h_m$.} Two choices for $h_m$ are studied: random sampling based and $k$-NN based. For this experiment, we fix the region numbers of the three IRL modules (from the shallow to the deep layers in our classification network) to 256, 128, and 64 respectively, and the point number $k$ per region for all the IRL modules to 6. Then we vary the sampled representative point number $m$ to be 1, 2, 4, and 6, to study the effects.
As shown in Table~\ref{table:sample_strategy}, for both sampling strategies, we can obtain strong results, especially for the $k$-NN based method which achieves the best performance with 81.0\% in OA. 
Moreover, we notice that either too small or too large value for the number of representative points $m$ will degrade the performance. The best classification result is obtained when we set $m$ to 4. We assume that too many points may contaminate the representation learning by introducing more redundant information, while fewer points are not sufficient to characterize the geometric information of a local region. 
Furthermore, we perform experiments by aggregating the point features in the local region. For simplicity, we use max-pooling (max) and mean-pooling (mean). 
However, both methods are worse than the $k$-NN based method. We hypothesize that the performance degradation can be attributed to the information loss caused by max-pooling or average-pooling, which only takes the maximum or average values of point features in each local region. 

\begin{table}[tb]
\small
\centering
\caption{Effectiveness of different components evaluated on the ScanObjectNN dataset~(\%)}
\begin{tabular}{|*{1}{p{1.0cm}<{\centering}}|*{3}{p{1.0cm}<{\centering}}|*{1}{p{1.0cm}<{\centering}}|*{1}{p{1.0cm}<{\centering}}|}
\hline
\multirow{2}{*}{Model}& \multicolumn{3}{c|}{ISL} & \multicolumn{1}{c|}{\multirow{2}{*}{IRL}} & \multirow{2}{*}{OA}\\
\cline{2-4}  & SFL & NFL & DFA & \multicolumn{1}{c|}{} & \multicolumn{1}{c|}{} \\
\hline
\hline
A & $\surd$ &  & &  & 66.5 \\
B &  & $\surd$ &  &  & 76.6\\
C & $\surd$ & $\surd$ &  &  & 78.7\\
D & $\surd$ & $\surd$ & $\surd$ &  & 79.8\\
E & $\surd$ & $\surd$ & $\surd$ & $\surd$ & \textbf{81.0} \\
\hline
\end{tabular}
\label{table:effectiveness_of_components}
\end{table}

\begin{table}[tb]
\small
\centering
\caption{Effectiveness of Region  Partition  Methods evaluated on the ScanObjectNN and KeypointNet dataset~(\%)}
\begin{tabular}{|*{1}{p{2.0cm}<{\centering}}|*{1}{p{2.0cm}<{\centering}}|*{2}{p{1.5cm}<{\centering}}|}
\hline
\multirow{2}{*}{Model}& \multicolumn{1}{c|}{ScanObjectNN} & \multicolumn{2}{c|}{\multirow{1}{*}{KeypointNet}}\\
\cline{2-4}  & OA & mIoU & mAP\\
\hline
\hline
FPS & 80.4 & 28.1 & 41.6\\
top-$S$ & 80.6 & 28.1 & 42.0\\
dilated top-$S$ & \textbf{81.0} & \textbf{28.6} & \textbf{43.2}\\
\hline
\end{tabular}
\label{table:effectiveness_of_different_region_partition}
\end{table}

\begin{table}[h] 
\small
\caption{The results~(\%) on the ScanObjectNN of different sampling strategies $h$ with varying number of representative points $m$ in each region, described in Eq.~\eqref{eq:h}. For the random sampling based method, we run the trained model 10 times on test set to report the results in the mean $\pm$ std format}
\centering
\begin{tabular}{|p{2.3cm}<{\centering}|p{1.3cm}<{\centering}|p{3.8cm}<{\centering}|}   
\hline
 $h$ & $m$ & OA  \\
\hline
\multirow{4}{*}{$k$-NN} & 1 & 80.3  \\
                    & 2  & 80.6 \\
                    & 4  & \textbf{81.0}\\
                    & 6  & 80.5\\
\hline
\multirow{4}{*}{random} & 1 & 80.17 $\pm$ 0.31\\
                    & 2   & 80.39 $\pm$ 0.35\\
                    & 4   & 80.52 $\pm$ 0.28 \\
                    & 6   & 80.44 $\pm$ 0.29 \\
\hline 
\multirow{1}{*}{mean} & - & 80.2 \\
\multirow{1}{*}{max} & -  & 80.4 \\
\hline
\end{tabular}

\label{table:sample_strategy}
\end{table}

\noindent\textbf{The Impact of Group Number.}\label{group_num}
To study the impact of the group number $S$, we first conduct experiments on ScanObjectNN by setting $S$ in the three IRL modules to be (256, 128, 64) as the base group numbers. Then we vary it in these modules at an equal rate of the base values. For simplicity, we adopt the $k$-NN based sampling strategy and change $S$ at the rate of $\frac{1}{4}$, $\frac{1}{2}$, 1, 2, 4. All the other parameters are fixed to the default values for fair comparisons. 
To give a more comprehensive analysis, results with $m=2$ and $m=4$ are reported. The detailed comparison results are summarized in Table~\ref{table:the impact of group numbers}. 
As shown, when we increase the rate of the base group numbers from $\frac{1}{4}$ to $1$ and set $m$ to 4, we can gain consistent improvement in OA. However, the performance drops when increasing the rate from $1$ to $4$. We find that too small and too large values for the group number are harmful. The best configurations for $S$ are (256, 128, 64) in the three different IRL layers, respectively. Similar phenomena can be observed when $m=2$. Specifically, the best performance is achieved when the rates are set to $1$ and $2$. The reason behind this may be that a small number of regions are insufficient to describe a 3D object, while a large number of regions may introduce redundant information and obstruct representation learning.    

\noindent\textbf{The visualization of the IRL module.} To further investigate our IRL module, we visualize how the other points influence the anchor points (red) in Fig.~\ref{Fig.ShapeNetPart Attention Visualization}. We observe that our IRL module can successfully capture different types of inter-region relations, such as symmetrical parts in space (the first three columns), semantic dependencies of different parts (the last three columns), which demonstrates the validity of our design logic.

\noindent\textbf{The Naive Version \emph{vs.} the Representative Point-based Version of the IRL module.}\label{complexity_comp}
To quantify the efficiency brought by the representative point-based version over the naive version of the IRL module, we measure their inference time~(ms). For fair comparisons, all the experiments are conducted on a platform with a single NVIDIA RTX 2080TI GPU. The batch size, the channel of the input feature, and point number $k$ per region are fixed to 32, 256, and 6, respectively. The detailed results are reported in Table~\ref{table:comparision}. As shown, compared with the naive version, the representative point-based method decreases the inference time by 67.8\%, when the point number $N$, the representative number $m$, and the group number $S$ are set to 1,024, 1, and 256. More significant improvements can be observed when we increase the input point number. For instance, when the point number reaches 4,096, the representative point-based version is around 7 times faster on inference. These results consistently demonstrate the efficiency of the representative point-based method. 


\begin{table}[!tb]
\small
\centering
\caption{The results~(\%) on the ScanObjectNN dataset by varying the group number $S$ and representative point number $m$ in each IRL module}
\begin{tabular}{|p{1.3cm}<{\centering}|p{3.8cm}<{\centering}|*{1}{p{2.3cm}<{\centering}}|}
\hline
$m$ & $S$ & OA \\
\hline
\hline 
\multirow{5}{*}{4} & (64, 32, 16) & 80.4\\
                    & (128, 64, 32) & 80.7\\
                    & (256, 128, 64) & \textbf{81.0} \\
                    & (512, 256, 128) & 80.7\\
                    & (1024, 512, 256) & 80.4\\
\hline 
\hline 
\multirow{5}{*}{2} & (64, 32, 16) & 80.4\\
                    & (128, 64, 32) &  80.5\\ 
                    & (256, 128, 64) &  80.6\\
                    & (512, 256, 128) & 80.6\\
                    & (1024, 512, 256) & 80.3\\
\hline

\end{tabular}

\label{table:the impact of group numbers}

\end{table}

\begin{table}[!tb]
\small
\centering
\caption{Comparison of inference time~(Inf.) between the naive and the representative point-based~(\emph{rep.}) methods. The lower the value, the better}
\begin{tabular}{|p{1.10 cm}<{\centering}|p{1.30cm}<{\centering}|*{1}{p{1.50cm}}<{\centering}|*{1}{p{1.3cm}}<{\centering}|*{1}{p{1.3cm}<{\centering}}|}

\hline
Points & Methods &$S$  & $m$  & Inf.~(ms)  \\
\hline
\hline 
\multirow{3}{*}{1,024} & \multirow{1}{*}{naive}       & 256     & - &  18.3 \\
\cline{2-5}          
                     & \multirow{2}{*}{\emph{rep.}}   & 256     & 1 &  5.9  \\
                     \cline{3-5}
                                                  &   & 256     & 4 &  8.9 \\
\hline
\multirow{3}{*}{2,048} & \multirow{1}{*}{naive}       & 512     & - &  53.3 \\
\cline{2-5}     
                      & \multirow{2}{*}{\emph{rep.}}  & 512     & 1 &  13.8 \\
		                     \cline{3-5}
                                          		  &   & 512     & 4 &  22.1 \\
\hline
\multirow{3}{*}{4,096} & \multirow{1}{*}{naive}       & 512     & - &  180.0 \\
\cline{2-5}
                     & \multirow{2}{*}{\emph{rep.}}   & 512     & 1 &  24.4 \\
		                     \cline{3-5}
		                                          &   & 512     & 4 &  33.2 \\
\hline
\end{tabular}

\label{table:comparision}

\end{table}

\section{Conclusion and Future Work}\label{sec:conclusion}
In this paper, we propose an end-to-end architecture named PRA-Net to exploit the intra-region contexts and inter-region relations for 3D point cloud analysis. 
Specifically, the ISL module extracts the intra-region contexts by dynamically integrating the local structural information into the point features, while the IRL module models the inter-region relations with the representative points adaptively sampled from each local region. 
Finally, two key components are combined in a principled way for deriving a more effective framework to learn highly discriminative representations for point clouds. 
Comprehensive experiments on four public benchmark datasets demonstrate the effectiveness and generality of our method. 
In the future, we would like to apply our framework to other point cloud analysis tasks such as 3D object detection and instance segmentation.



\ifCLASSOPTIONcaptionsoff
  \newpage
\fi



%
{\small
\bibliographystyle{IEEEtran}
\bibliography{references}
}

%




\begin{IEEEbiography}[{\includegraphics[width=1in,height=1.25in,clip,keepaspectratio]{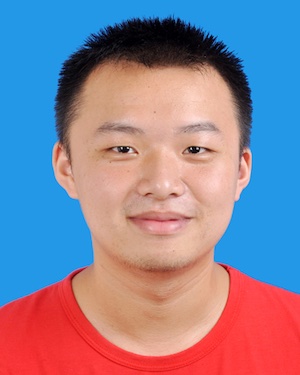}}]{Silin Cheng}
received his B.S. degree from the School of Computer Science and Technology, Huazhong University of Science and Technology (HUST), China in 2019. He is currently a master
student with the School of Electronic Information and Communications, HUST. His main research interests include 3D shape analysis, 3D scene understanding and continue learning.
\end{IEEEbiography}

\begin{IEEEbiography}[{\includegraphics[width=1in,height=1.25in,clip,keepaspectratio]{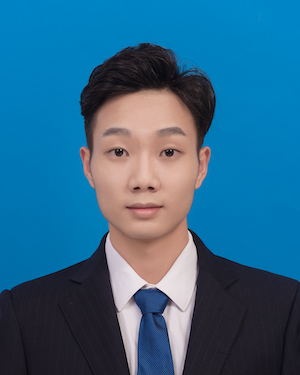}}]{Xiwu Chen}
received his B.S. degree from the School of Computer Science and Technology, Huazhong University of Science and Technology (HUST), China in 2019. He is currently a master student with the School of Electronic Information and Communications, HUST. His main research interests include 3D shape analysis and 3D object detection.
\end{IEEEbiography}

\begin{IEEEbiography}[{\includegraphics[width=1in,height=1.25in,clip,keepaspectratio]{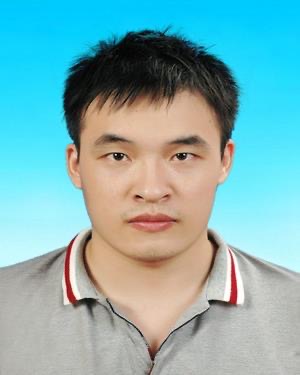}}]{Xinwei He}
received his Ph.D. degree in Electronics and Information Engineering from Huazhong University
of Science and Technology (HUST), Wuhan, China. His research interests include image caption, 3D shape analysis and 3D object detection.
\end{IEEEbiography}

\begin{IEEEbiography}[{\includegraphics[width=1in,height=1.25in,clip,keepaspectratio]{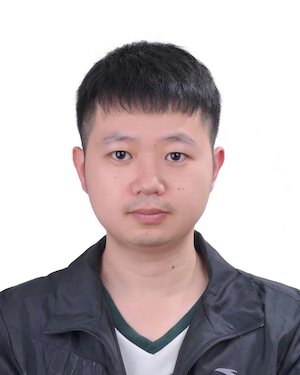}}]{Zhe Liu}
received his B.S. from Hubei University of Technology (HBUT) and M.S. from Huazhong University of Science and Technology (HUST). He is currently studying for his doctor in HUST. He has published 3 papers in the areas of computer vision such as ECCV, AAAI. He also served as a reviewer for IJCAI-2021 and AAAI-2021. His current research interests include pattern recognition, deep learning, 3D computer vision.
\end{IEEEbiography}

\begin{IEEEbiography}[{\includegraphics[width=1in,height=1.25in,clip,keepaspectratio]{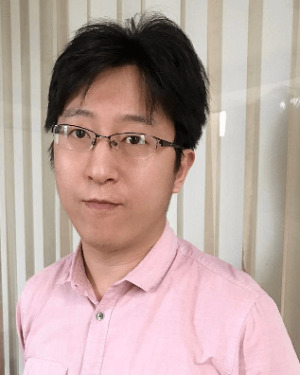}}]{Xiang Bai} received his B.S., M.S., and Ph.D. degrees from the Huazhong University of Science
and Technology (HUST), Wuhan, China, in 2003,
2005, and 2009, respectively, all in electronics and
information engineering. He is currently a Professor
with the School of Artificial Intelligence and Automation, HUST. His research interests include
object recognition, shape analysis, and OCR. He
has published more than 150 research papers. He
is an editorial member of IEEE TPAMI, Pattern
Recognition, and Frontier of Computer Science. He
is the recipient of 2019 IAPR/ICDAR Young Investigator Award for his outstanding contributions to scene text understanding. He is a senior member of IEEE.
\end{IEEEbiography}


\vfill


\end{document}